\documentclass{article}

    \PassOptionsToPackage{numbers, compress}{natbib}

\usepackage[final]{neurips_2019}

\usepackage[utf8]{inputenc} 
\usepackage[T1]{fontenc}    
\usepackage{hyperref}       
\usepackage{url}            
\usepackage{booktabs}       
\usepackage{amsmath}
\usepackage{amsfonts}       
\usepackage{nicefrac}       
\usepackage{microtype}      
\usepackage{bbm}
\usepackage{color}
\usepackage{multirow}
\usepackage{graphicx}
\usepackage{caption}
\usepackage{subcaption}
\usepackage{amssymb}
\usepackage{enumitem}
\usepackage{makecell}
\usepackage[symbol]{footmisc}
\usepackage{float}
\usepackage{mwe,tikz}
\usepackage[percent]{overpic}
\usepackage{algorithm}
\usepackage[noend]{algpseudocode}
\usepackage{tablefootnote}
\usepackage{tabu}

\algnewcommand{\LineComment}[1]{\State \(\triangleright\) #1}

\newcommand\indicator[1]{\mathbbm{1}[#1]}

\title{General Evaluation for Instruction Conditioned Navigation using Dynamic Time Warping}

  \author{Gabriel Ilharco$^\dagger$\thanks{$\ $ Work done as a member of the Google AI Residency Program.} \quad Vihan Jain$^{\ddagger}$ \quad Alexander Ku$^{\ddagger}$ \quad Eugene Ie$^{\ddagger}$ \quad Jason Baldridge$^{\ddagger}$\\
  $^\dagger$Paul G. Allen School of Computer Science \& Engineering, 
  University of Washington \\
  $^\ddagger$Google Research \\
  {\tt gamaga@cs.washington.edu}, {\tt \{vihanjain,alexku,eugeneie,jridge\}@google.com} \\}

\begin{document}

\maketitle

\begin{abstract}
In instruction conditioned navigation, agents interpret natural language and their surroundings to navigate in an environment. Datasets for such tasks typically contain pairs of these instructions and reference trajectories, but current popular evaluation metrics fail to properly account for the fidelity of agents to the those trajectories. To address this, we introduce the normalized Dynamic Time Warping (nDTW) metric. nDTW softly penalizes deviations from the reference path, is naturally sensitive to the order of the nodes composing each path, is suited for both continuous and graph-based evaluations, and can be efficiently calculated. Further, we define SDTW, which constrains nDTW to only successful episodes and effectively captures both success and fidelity. We collect human similarity judgments for simulated paths and find our DTW metrics correlates better with human rankings than all other metrics. We also show that using nDTW as a reward signal for agents using reinforcement learning improves performance on both the Room-to-Room and Room-for-Room datasets.


\end{abstract}
\section{Introduction}

Following natural language instructions is essential for flexible and intuitive interaction between humans and embodied agents \cite{Chen:2011:LIN}. Recent advances in machine learning and high availability of processing power has greatly lessened some of the technical barriers for learning these behaviours. In particular, there has been growing interest in the task of Vision-and-Language Navigation (VLN), where agents use language instructions and visual stimuli to navigate in a virtual---sometimes photo-realistic---\cite{misra-etal-2018-mapping,fu2018from,anderson2018vision,rerere,Chen19:touchdown} or physical environment \cite{kyriacou2005vision,thomason2015learning,WilliamsGRT18}. For these language conditioned behaviors, evaluation metrics should be sensitive to both the task itself and to how the instructions informs the task.

\noindent

The most obvious way of evaluating agents in goal-oriented tasks is whether they reach their goals (denoted as \emph{Success Rate} (SR)). However, this measure has its drawbacks when applied to path-oriented tasks \cite{jain2019stay}. \textit{Success weighted by Path Length} (SPL), introduced by  \citet{Anderson:2018:Evaluation}, is an evaluation metric that rewards agents that reached their goals and do so efficiently with respect to the length of their trajectory. Since then, new metrics like \textit{Success weighted by Edit Distance} (SED) \cite{Chen19:touchdown} and \textit{Coverage weighted by Length Score} (CLS) \cite{jain2019stay} have been proposed that additionally take into account intermediary states along the agent's trajectory. The analysis performed in \citet{jain2019stay} (as well as in Appendix of this work) reveal several limitations of these metrics resulting in a gap in the evaluation of such agents.

\begin{figure*}
\centering
\includegraphics[width=.8\linewidth]{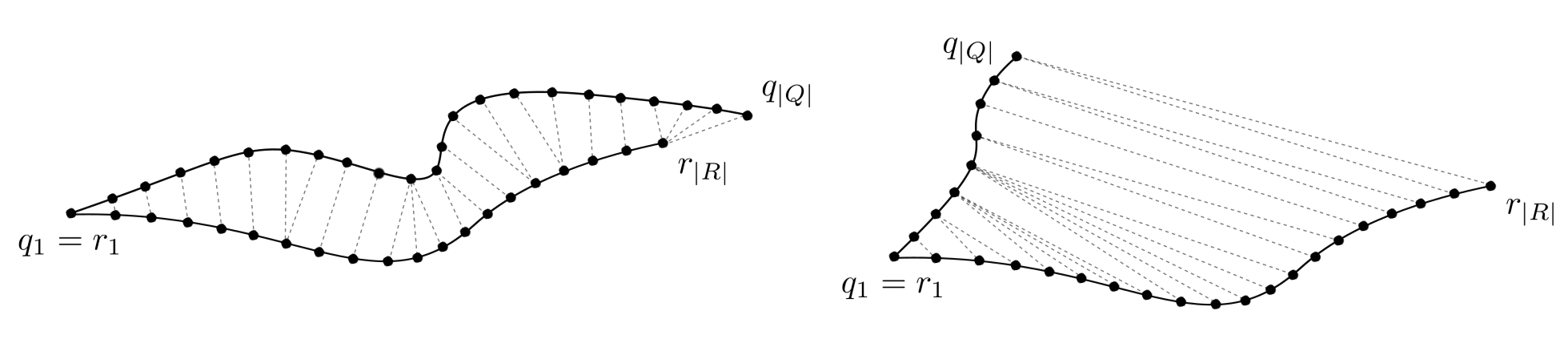}
\caption{Illustration of two pairs of reference ($R=r_{1..|R|}$) and query ($Q=q_{1..|Q|}$) series (solid), and the optimal warping between them (dashed) when computing DTW.}
\label{fig:dtw_illustration}
\end{figure*}

Dynamic Time Warping (DTW) \cite{berndt1994using} is a similarity function for time-series which is long used in speech processing \cite{myers1980performance, sakoe1990dynamic, muda2010voice}, robotics \cite{schmill1999learned, vakanski2012trajectory}, data mining \cite{keogh2000scaling, rakthanmanon2012searching}, handwriting recognition \cite{rath2003word}, gesture recognition \cite{ten2007multi, akl2010accelerometer} and more \cite{legrand2008chromosome, rebbapragada2009finding, keogh2009supporting}. DTW identifies an optimal warping---an alignment of elements from a reference and a query series such that the cumulative distance between aligned elements is minimized, as illustrated in Figure \ref{fig:dtw_illustration}. In this work, we adapt the DTW measure for evaluating instruction conditioned navigation agents, demonstrating its strengths against previously proposed metrics and further showing that it can be used as a reward signal to improve training.

\section{Dynamic Time Warping for Navigation}
\label{sec:dtw}

DTW is computed by aligning elements of a reference $R$ and a query $Q$ series while preserving their order. If the elements of the two series belong to some feature space $\mathcal{F}$ (in the context of navigation, $\mathcal{F}$ is the space of navigable points in space), then DTW finds the optimal ordered alignment between the two series by minimizing the cumulative cost:

\begin{equation}
    \text{DTW}(R, Q) = \min_{W \in \mathcal{W}} \sum_{(i_k, j_k) \in W} \delta(r_{i_k}, q_{j_k}).
\end{equation}

\noindent
where $\delta:\mathcal{F}\times\mathcal{F} \rightarrow \mathbb{R}_{\ge 0}$ is some distance function mapping pairs of elements from the two series to a real non-negative number and $W = w_{1..|W|}$ is a \textit{warping} with $w_k=(i_k,j_k) \in [1:|R|] \times [1:|Q|]$, respecting the step-size ($w_{k+1} - w_k \in {(1, 1), (1, 0), (0, 1)}$) and boundaries ($w_1 = (1,1)$ and $w_{|W|} = (|R|, |Q|)$) conditions.

DTW can be adapted to the context of navigation in discrete environments by using the shortest distance along the environment graph from node $r_i$ and $q_j$ as the cost function. For some applications, the cost function can be the Euclidean distance between the coordinates of any two points in space. In general continuous environments where obstruction can be an issue, one can pre-compute pairwise distances from fixed grid points, and approximate the distance at runtime by finding the closest grid points to $r_i$ and $q_j$. 

The ideal metric for evaluating navigation agents must not be sensitive to the scale and density of nodes along the agent trajectories so that the metric is comparable between different applications (e.g., indoor \cite{anderson2018vision} and outdoor \cite{Chen19:touchdown,cirik2018following,streetlearn}). 
We adapt DTW---a sum of at least $|R|$ distance terms---by normalizing it by a factor of $\frac{1}{|R|\cdot d_{th}}$, where $d_{th}$ is a sampling rate invariant threshold distance defined for measuring success. Further, to aid interpretability, we take the negative exponential of this normalized value, resulting in a score bounded between 0 and 1, yielding higher numbers for better performance. In summary, \textit{normalized Dynamic Time Warping} (nDTW) is composed by these operations sequentially applied to DTW, as shown in Eq. \ref{eq:ndtw}.
\begin{equation}
    \textrm{nDTW}(R, Q) = \exp\left(-\dfrac{\textrm{DTW}(R, Q)}{|R|\cdot d_{th}}\right) = \exp\left({-\dfrac{\min\limits_{W \in \mathcal{W}} \sum_{(i_k, j_k) \in W} d(r_{i_k}, q_{j_k})}{|R|\cdot d_{th}}}\right)
    \label{eq:ndtw}
\end{equation}
Since it is directly derived from DTW, nDTW can be exactly computed in quadratic time and space complexity  and approximately computed in linear time and space complexity, as described in the Appendix. In algorithms such as FastDTW \cite{salvador2007toward} where interpolation is required, a simple adaptation---sampling nodes---can be made for discrete environments.

\textit{nDTW} has many desirable properties for scoring agent paths. It measures similarity between the \textit{entirety} of two trajectories, softly penalizing deviations. It naturally captures the importance of the goal by forcing the alignment between the final nodes of the trajectories. It is insensitive to changes in scale and density of nodes, but sensitive to the order in which nodes compose trajectories.. It can be used for continuous path evaluation \cite{Blukis:18visit-predict} as well as graph-based evaluation. The optimal warping can be computed exactly in quadratic time and approximately in linear time. Finally, it is more generally applicable to any task requiring matching a sequence of actions provided an element-wise distance function is available.


Due to the popularity of metrics like SPL \cite{Anderson:2018:Evaluation} and SED \cite{Chen19:touchdown} that stress the importance of reaching the goal in navigation tasks \cite{anderson2018vision,Chen19:touchdown}, we analogously define \textit{Success weighted by normalized Dynamic Time Warping} (SDTW), given by $\textrm{SDTW}(R, Q) = \textrm{SR}(R, Q) \cdot \textrm{nDTW}(R, Q)$, where $\textrm{SR}(R, Q)$ is one if the episode was successful and zero otherwise, commonly defined by the threshold distance $d_{th}$.

\section{Evaluation}
\label{sec:discussion}

To assess the utility of nDTW as a measure of similarity between two paths, we compare its correlation with human judgments for simulated paths in the context of other standard metrics. Further, we demonstrate that it is advantageous to use it as a reward signal for RL agents on the Room-to-Room (R2R) task \cite{anderson2018vision} and its Room-for-Room (R4R) extension \cite{jain2019stay}.

\subsection{Human evaluations}

\begin{figure*}
\centering
\includegraphics[width=.8\linewidth]{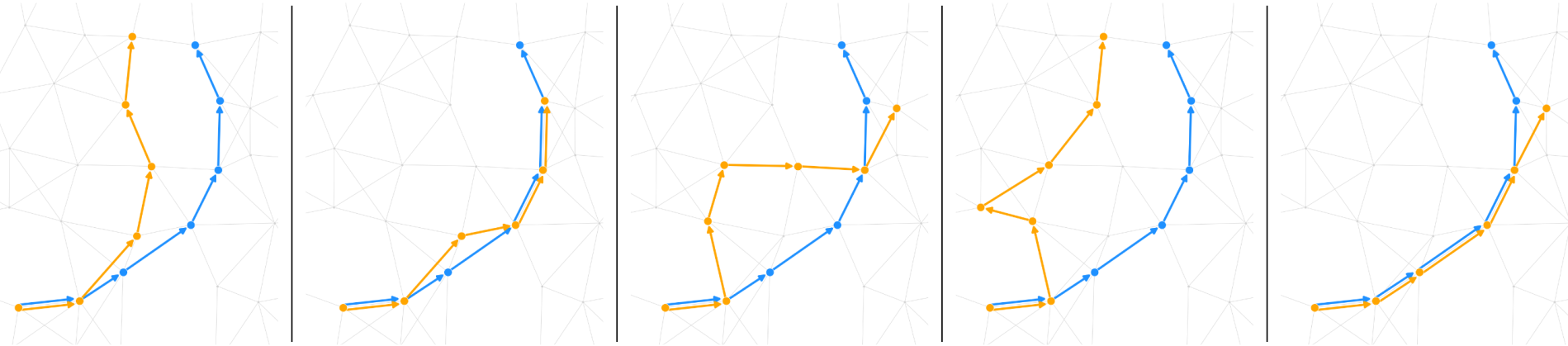}
\caption{Example comparison set with one reference path (blue) and five query paths (orange).}
\label{img:human_evals}
\end{figure*}

As illustrated in Figure \ref{img:human_evals}, we give human raters a series of questionnaires each containing a set of five reference (shown in blue) and query (shown in orange) path pairs. In each set, we keep the reference path fixed and instruct raters to rank the path pairs in response to this question: \textit{``If I instructed a robot to take the blue path, which of these orange paths would I prefer it to take?"}

The environment and paths are randomly generated. The environment consists of $15{\times}15$ nodes, forming an approximate grid. Each node $n_{ij}$ has coordinates $(x_{ij}, y_{ij})$, where $x_{ij} \sim U(i-\zeta, i+\zeta)$ and $y_{ij} \sim U(j-\zeta, j+\zeta)$ are independently drawn according to a parameter $\zeta$ (set to $0.3$). For every pair of nodes $n_{i_1j_1}$ and $n_{i_2j_2}$, we take the Euclidean distance between their coordinates $\lVert n_{i_1j_1}  n_{i_2j_2} \rVert$ and connect them with an edge if and only if $\lVert n_{i_1j_1}  n_{i_2j_2} \rVert \le 1.4$. Each path is generated according to a random procedure: first, a random node is drawn; then, a node two or three edges away is chosen, and this step is repeated. The final path is the shortest path connecting adjacent nodes from this procedure. As in \citet{anderson2018vision}, we set the success threshold to be $1.33$ times the average edge length in the environment.

\begin{table*}
    \centering
    \setlength\tabcolsep{2.5pt}
    \begin{tabular}{ccccccccccc}
                    & \multicolumn{6}{c}{\textbf{UC}  (\textit{nDTW vs})}          & \multicolumn{4}{c}{\textbf{SC}  (\textit{SDTW vs})} \\\cmidrule(lr{0.2cm}){2-7}\cmidrule{8-11}
                    &  PL & NE & ONE  & CLS & AD & MD & SR & OSR & SPL & SED \\\hline
          +/-       & 242/17 & 254/9 & 255/9 & 162/46 & 254/12 & 253/12 & 219/16 & 220/14 & 219/17 & 213/26  \\
          sign test & 4.1e-52  & 2.0e-63 & 1.0e-63 & 2.4e-16   & 6.9e-60  & 6.9e-60 & 9.6e-47  & 8.8e-49  & 6.7e-46  & 1.1e-37 \\
    \end{tabular}
    \caption{Binomial tests on how different metrics compare in correlation with human judgments. The sign test uses \emph{n}=sum of positives and negatives; \emph{k}=number of positives; \emph{p}=0.5.}
    \label{tab:human_evals}
\end{table*}

We conduct two studies: \emph{unconstrained} (UC) where we compare nDTW with popular continuous metrics like path length (PL), navigation error (NE), oracle navigation error (ONE), CLS \cite{jain2019stay}, average deviation (AD), maximum deviation (MD); and \emph{success-constrained} (SC) where we compare SDTW with metrics that that are non-zero only if a success criteria is met---namely, success rate (SR), oracle success rate (OSR), SPL \cite{Anderson:2018:Evaluation}, and SED \cite{Chen19:touchdown}. These metrics are discussed in detail in the Appendix. We collect annotations on 2525 samples (505 sets of 5 query and reference pairs) from 9 human raters, split between UC (1325 samples) and SC (1200 samples).

To analyze the collected annotations, we first assign nDTW in UC study (similarly SDTW in SC study) a positive/negative sign depending if it has higher/lower correlation than competing metric for a given human ranking of query paths with respect to a reference path, and then compare across all reference paths (discarding ties) using a sign test. Table \ref{tab:human_evals} depicts the results of the binomial test for the null hypothesis. Both nDTW and SDTW correlate substantially better with human orderings, compared to the competing metrics in their respective categories.

\begin{table*}
\centering
\setlength\tabcolsep{2.9pt}
\begin{tabular}{lcccccccccccc}
  & \multicolumn{6}{c}{\textbf{R2R}} & \multicolumn{6}{c}{\textbf{R4R}}\\\cmidrule(lr{0.2cm}){2-7}\cmidrule{8-13}
Agent                            & SR & SPL & SED & CLS & \underline{nDTW} & \underline{SDTW} &  SR & SPL & SED & CLS & \underline{nDTW} & \underline{SDTW}\\\hline\noalign{\vskip .02in}
random & 5.1 & 3.3 & 5.8 & 29.0 & 27.9 & 3.6 & 13.7 & 2.2 & 16.5 & 22.3 & 18.5 & 4.1 \\
goal-oriented & 43.7 & 38.4 & 31.9 & 53.5 & 54.4 & 36.1 & \textbf{28.7} & 15.0 & \textbf{9.6} & 33.4 & 26.9 & 11.4 \\
fidelity-oriented & \textbf{44.4} & \textbf{41.4} & \textbf{33.9} & \textbf{57.5} & \textbf{58.3} & \textbf{38.3} & 28.5 & \textbf{21.4} & 9.4 & \textbf{35.4} & \textbf{30.4} & \textbf{12.6} \\
\end{tabular}
\caption{Evaluation metrics as percentages on R2R and R4R Validation Unseen sets for agents with different reward functions. In all metrics, higher means better.\label{tab:rxr}}
\end{table*}

\subsection{Evaluation on VLN Tasks}

We demonstrate a practical application of nDTW, by using it as a reward signal for agents in the Matterport3D environment \cite{Matterport3D}, on both the R2R \cite{anderson2018vision} and R4R datasets \cite{jain2019stay}. Unlike R2R, which contains only direct-to-goal reference paths, R4R contains more convoluted paths that might even come back to their starting point. In the latter scenario, the overly simplistic nature of previous metrics clearer. We follow the experimental settings of \citet{jain2019stay}, and train agents using our own implementation. A random baseline as in \cite{jain2019stay} is computed by sampling the step count from the distribution of reference path lengths in the datasets. Each step is taken by uniformly sampling between possible neighbors and scores are averaged over one million of these random trajectories.

Our \textit{goal-oriented} agent receives at each transition $q_i \rightarrow q_{i+1}$ a reward equal to how much closer it got to the goal $g$: $d(q_{i}, g) - d(q_{i+1}, g)$. At the end of the episode $q_f$, the agent receives a completion reward of +1 if it was deemed successful (i.e., it stopped within an acceptable distance from goal) and -1 otherwise. Our \textit{fidelity-oriented} agent receives at each transition a reward proportional to the gain in nDTW score with respect to the reference path $R$: $\textrm{nDTW}(q_{1..i+1}, R) - \textrm{nDTW}(q_{1..i}, R)$. At the end of the episode, the agent receives a non-zero reward that is a linear function of its navigation error: $1 - d(q_f, g) / d_{th}$ only if it was successful ($d_{th} = 3m$ in Matterport3D environment).


The metrics for the random, goal-oriented and fidelity-oriented agents are shown in Table \ref{tab:rxr}. Compared to a goal-oriented reward strategy, taking advantage of nDTW as a reward signal not only results in better performance on nDTW and SDTW metrics but also better performance on prior metrics like CLS and SPL. nDTW shows better differentiation compared to CLS on R4R between goal and fidelity oriented agents. SED scores random paths more highly than those of trained agents, and neither SR nor SED differentiate between goal and fidelity orientation. SPL appears to do so (15.0 vs 21.4), but this is only due to the fact that the fidelity-oriented agent produces paths that have more similar length to the reference paths rather than fidelity to them. As such, SDTW provides the clearest signal for indicating both success and fidelity.

\section{Conclusion}
\label{sec:conclusion}

In this work, we adapt DTW to the context of instruction conditioned navigation to introduce a metric that does not suffer from shortcomings of previous evaluation metrics. The many desirable properties of our proposed metric for evaluating path similarity, nDTW, are reflected both qualitatively in human evaluations---which prefer nDTW over other metrics---and practically in VLN agents---that see performance improvements when using nDTW as a reward signal. For assessing performance of instruction conditioned navigational agents, our proposed SDTW captures well not only the success criteria of the task, but also the similarity between the intended and observed trajectory. While multiple measures (especially path length and navigation error) are useful for understanding different aspects of agent behavior, we hope the community will adopt SDTW as a single summary measure for future work and leaderboard rankings.

\bibliographystyle{plainnat}
\bibliography{example}  

\newpage

\section*{Appendix}

\appendix

\section{Evaluation metrics in instruction conditioned navigation}

 Let $\mathcal{P}$ be the space of possible paths, where each $P \in \mathcal{P}$ is a sequence of observations $p_{1..|P|}$. An evaluation metric that measures the similarity between two paths is then some function $f : \mathcal{P} \times \mathcal{P} \rightarrow \mathbb{R}$, where $f(Q, R)$ maps a query path $Q$ and a reference path $R$ to a real number. We denote by $d(n, m)$ the distance of the shortest path between two nodes $n$ and $m$, and $d(n, P) = \min_{p \in P} d(n, p)$ the shortest distance between a node and a path. In discrete scenarios, $d(n, m)$ can be exactly computed using Dijkstra's algorithm \cite{dijkstra1959note}. In continuous scenarios, one strategy for computing $d(n, m)$ is to divide the environment into a grid of points so that they are at most some error margin of each other. The distance $d(n, m)$ between all pairs of grid points can be efficiently pre-computed \cite{tsitsiklis1995efficient}, and the distance between any pair of points can then be obtained within some error margin. In environments where there are no obstacles, $d(n, m)$ can be computed in constant time by taking the Euclidean distance between the points. Commonly, a threshold distance $d_{th}$ is defined for measuring success. Table \ref{tab:nav-metrics} defines existing and proposed metrics for instruction conditioned navigation. All previously proposed metrics fall short in different ways.


Of the existing metrics for assessing performance in instruction conditioned navigation, the majority are not intended to measure fidelity between two paths $Q$ and $R$: \textit{Path Length} (PL) measures the length of the query path, optimally equal to the length of the reference path; \textit{Navigation Error} (NE) measures the distance between the last nodes of the query and reference paths; \textit{Oracle Navigation Error} (ONE) measures the distance from the last node in the reference path to the query path; \textit{Success Rate} (SR) measures whether the last node in the predicted path is within $d_{th}$ of the last node in the reference path; \textit{Oracle Success Rate} (OSR) measures whether the distance between the last node in the reference path and the query path is within $d_{th}$; finally, \textit{Success weighted by Path Length} (SPL) \cite{Anderson:2018:Evaluation} weights SR with a normalized path length. None of these metrics take into account the entirety of the reference path $R$ and thus are less than ideal for measuring similarity between two paths. Because they are only sensitive to the the last node in the reference path, these metrics are tolerant to intermediary deviations. As such, they mask unwanted and potentially dangerous behaviours in tasks where following the desired action sequence precisely is crucial \cite{jain2019stay}.

\textit{Success weighted by Edit Distance} (SED)~\cite{Chen19:touchdown} uses the Levenshtein edit distance $\text{ED}(R, Q)$ between the two action sequences $A_R = ((r_1, r_2), (r_2, r_3), ..., (r_{|R|-1}, r_{|R|}))$ and $A_Q = ((q_1, q_2), (q_2, q_3), ..., (q_{|Q|-1}, q_{|Q|}))$. When computing $\text{ED}(R, Q)$, SED does not take into account the distance between path components, but only checks if the actions are a precise match or not. This shortcoming becomes clear in continuous or quasi-continuous scenarios: an agent that travels extremely close to the reference path---but not exactly on it---is severely penalized by SED.

\begin{table*}
\centering
\setlength\tabcolsep{4.8pt}
\begin{tabular}{lcl}

Metric     & $\uparrow$ $\downarrow$ & Definition\\\hline\noalign{\vskip .05in}  
Path Length (PL)              &      -       & $\sum_{1 \le i < |Q|} d(q_i, q_{i+1})$\\
Navigation Error (NE)         & $\downarrow$ & $d(q_{|Q|}, r_{|R|})$\\
Oracle Navigation Error (ONE) & $\downarrow$ & $\min_{q \in Q}d(q, r_{|R|})$\\
Success Rate (SR)             & $\uparrow$   & $\indicator{\text{NE}(R, Q) \le d_{th}}$\\
Oracle Success Rate (OSR)     & $\uparrow$   & $\indicator{\text{ONE}(R, Q) \le d_{th}}$\\ 
Average Deviation (AD)      & $\downarrow$   & $\sum_{q \in Q}d(q, R)/|Q|$\\
Max Deviation (MD)      & $\downarrow$   & $\max_{q \in Q}d(q, R)$\\
Success weighted by PL (SPL)  &  $\uparrow$   & $\text{SR}(R, Q) \cdot \frac{d(q_1, r_{|R|})}{\max\{\text{PL}(Q), d(q_1, r_{|R|})\}}$\\
Success weighted by Edit Distance (SED)         & $\uparrow$ & $\text{SR}(R, Q) \cdot \left(1 - \frac{\text{ED}(A_R, A_Q)}{\max{\{|A_R|, |A_Q|}\}}\right)$\\
Coverage weighted by Length Score (CLS)      & $\uparrow$   & $\text{PC}(R, Q) \cdot \text{LS}(R, Q)$\\
\hline\noalign{\vskip .05in}    
Normalized Dynamic Time Warping (nDTW) & $\uparrow$ & $\exp\left({-\frac{\min\limits_{W \in \mathcal{W}} \sum_{(i_k, j_k) \in W} d(r_{i_k}, q_{j_k})}{|R|\cdot d_{th}}}\right)$\\
Success weighted by nDTW (SDTW) & $\uparrow$ & $\textrm{SR}(R, Q)\cdot \textrm{nDTW}(R, Q)$

\\\hline
\end{tabular}
\caption{Metrics and optimal directions, which agents should minimize ($\downarrow$) or maximize ($\uparrow$). \label{tab:nav-metrics}}
\end{table*}

\textit{Coverage weighted by Length Score} (CLS) \cite{jain2019stay} computes the path coverage $\text{PC}(R, Q) = \frac{1}{|R|} \sum_{r \in R}\exp\left(-\frac{d(r, Q)}{d_{th}}\right)$ and a length score $\text{LS}(R, Q) = \frac{\text{PC}(R, Q) \cdot \text{PL}(R)}{\text{PC}(R, Q) \cdot \text{PL}(R) + |\text{PC}(R, Q) \cdot \text{PL}(R) - \text{PL}(Q)|}$, combining them by multiplication. Although it addresses the major concerns of previous metrics, CLS is not ideal in some scenarios. For instance, because $\text{PC}(R, Q)$ is order-invariant, for a given reference path $R = (a, b, c, a)$, an agent that navigates $Q_1{=}(a, c, b, a)$ and one that executes a trajectory $Q_2{=}(a, b, c, a)$ both have the same CLS score. If an instruction such as \textit{``Pick up the newspaper in the front door, leave it in my bedroom and come back"} is given, an agent that navigates along the intended path in the reverse order would be incapable of completing its task.

We include two additional simple metrics that capture a single summary measure of the difference between two paths. \textit{Average Deviation} (AD) and \textit{Max Deviation} (MD) measure the average and maximum deviations from points on the query path, with respect to the entire reference path. Although these metrics take into account both paths in their totality and measure similarity to some extent, they are critically flawed by not taking into account the order of the nodes.

\section{Implementing of proposed metrics}  

\begin{algorithm}
\caption{nDTW}
\begin{algorithmic} 
\Require{$R=r_{1..|R|}$ (reference path); $Q=q_{1..|Q|}$ (query path); $d_{th}$ (success distance threshold).}
\Ensure{The nDTW score.}
\Procedure{nDTW}{$R$, $Q$, $d_{th}$}

\State $\textbf{C}$ = array([$|R|+1$, $|Q|+1$], default=inf)
\State $\textbf{C}[0][0] \leftarrow$ 0

\State
\For{$i \leftarrow 1$ to $|R|$}
    \For{$j \leftarrow 1$ to $|Q|$}
        \State $\textbf{C}[i][j] \leftarrow \delta(R[i], Q[j]) + \min_{(p, q) \in \{(i-1,j), (i,j-1), (i-1,j-1)\}}\textbf{C}[p][q]$
    \EndFor
\EndFor

\State
\State
\Return math.exp(-$\textbf{C}[|R|][|Q|]$ / ($|R|\cdot d_{th}$))
\EndProcedure
\end{algorithmic}
\label{alg:ndtw}
\end{algorithm}

\begin{algorithm}
\caption{SDTW}
\begin{algorithmic} 
\Require{$R=r_{1..|R|}$ (reference path); $Q=q_{1..|Q|}$ (query path); $d_{th}$ (success distance threshold).}
\Ensure{The SDTW score.}
\Procedure{SDTW}{$R$, $Q$, $d_{th}$}
\If{$\delta(R[i], Q[j]) > d_{th}$}
    \State
    \Return 0.0
\EndIf
\State
\State
\Return nDTW($R$, $Q$, $d_{th}$)
\EndProcedure
\end{algorithmic}
\label{alg:sdtw}
\end{algorithm}

To compute nDTW, we can define a matrix $\textbf{C} \in \mathbb{R}^{(|R|+1) \times (|Q|+1)}$ where 
\begin{equation}
\textbf{C}_{i,j} := \text{DTW}(r_{1..i}, q_{1..j}) 
\end{equation}
\noindent
for $(i, j) \in [0:|R|] \times [0:|Q|]$. All elements in this matrix can be computed in $O(|R||Q|)$, using dynamic programming, as shown in Algorithm \ref{alg:ndtw}. The key to do so is realizing that $\textbf{C}_{i,j}$ depends only on $\textbf{C}_{i-1,j}$, $\textbf{C}_{i,j-1}$ and $\textbf{C}_{i-1,j-1}$. Therefore, we can efficiently compute $\textrm{DTW}(R,Q) = \textbf{C}_{|R||Q|}$ by filling out the slots in matrix $\textbf{C}$ in an ordered fashion: rows are sequentially computed from 1 to $|R|$ and, in each of them, columns are computed from 1 to $|Q|$. Note that this allows constant time computing of each $\textbf{C}_{i,j}$, since $\textbf{C}_{i-1,j}$, $\textbf{C}_{i,j-1}$ and $\textbf{C}_{i-1,j-1}$ are previously computed. As initial conditions, $\textbf{C}_{0,0} = 0$ and $\textbf{C}_{i,0} = \inf$ and $\textbf{C}_{0,j} = \inf$, for $1 \le i \le |R|$ and $1 \le j \le |Q|$. nDTW can be computed by applying the normalizing operations to $\textbf{C}[|R|][|Q|]$ and SDTW, as shown in Algorithm \ref{alg:sdtw}, limits nDTW to only successful episodes.

DTW, and consequently nDTW and SDTW can be approximated in linear time and space complexity. We refer readers to \citet{salvador2007toward} for explanations, proof and pseudo-code. In scenarios where long paths are common, this computational efficiency affords the opportunity to apply nDTW as a evaluation function as well as a reward signal for reinforcement learning agents.

\section{Visualizing nDTW scores}

\begin{figure*}
\centering
\includegraphics[width=\linewidth]{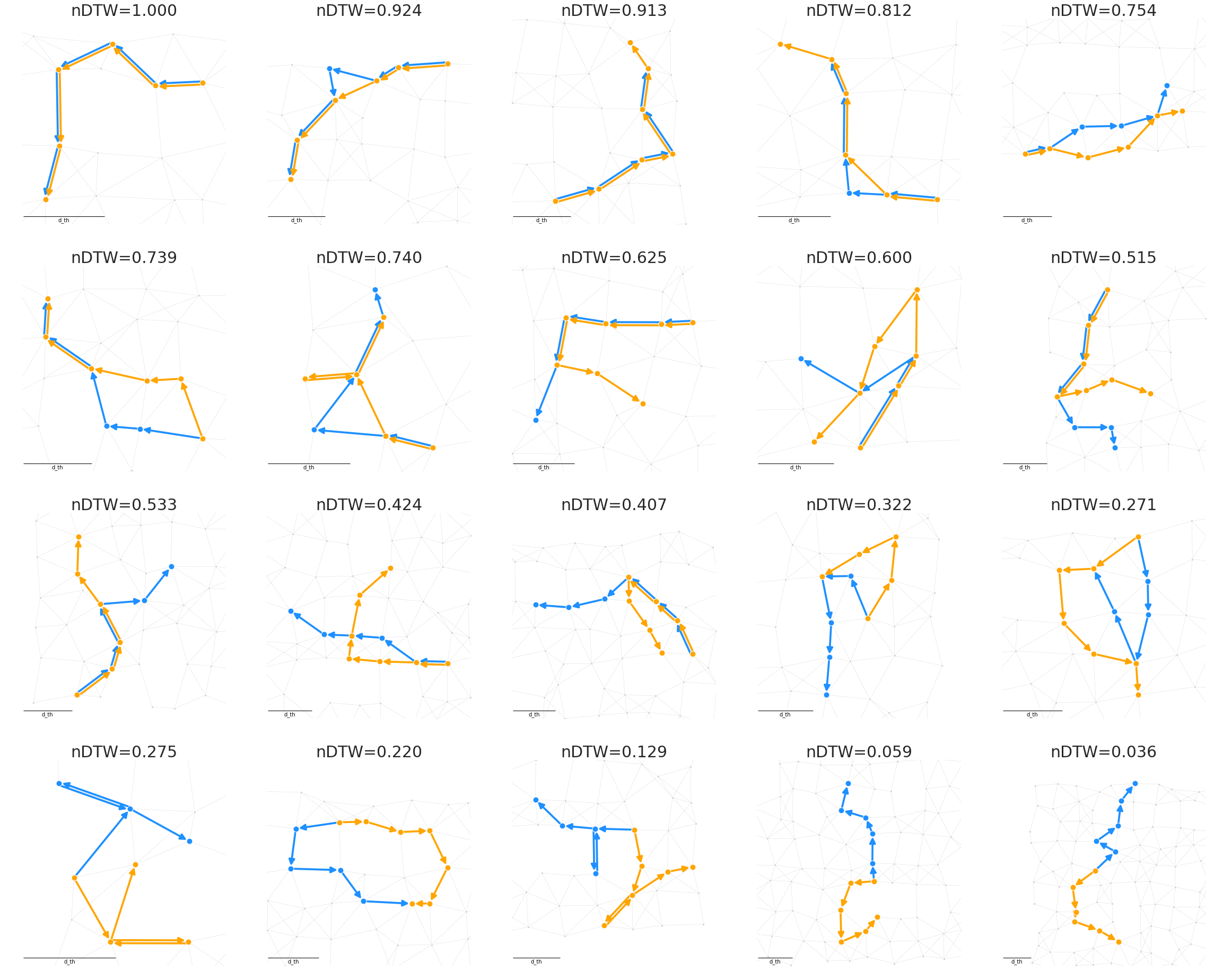}
\caption{Examples of random reference (blue) and query (orange) paths, sorted by nDTW values.}
\label{img:dtw}
\end{figure*}

Figure \ref{img:dtw} illustrates multiple pairs of reference (blue) and query (orange) paths, accompanied (and sorted) by their nDTW values.

\end{document}